\documentclass{article}
\usepackage{float}
\usepackage{graphicx}
\usepackage{multicol}
\usepackage{authblk}

\graphicspath{{figures}}

\title{A Self-Supervised Approach to Land Cover Segmentation}
\author[1,*]{Charles Moore}
\author[2,3,*]{Dakota Hester}
\affil[1]{\small Department of Computer Science and Engineering, Mississippi State University, Starkvile, MS 39759, USA}
\affil[2]{Department of Sustainable Bioproducts, Mississippi State University, Starkvile, MS 39759, USA}
\affil[3]{Department of Agricultural and Biological Engineering, Mississippi State University, Starkvile, MS 39759, USA}
\affil[*]{Corresponding authors: \{cam1271,dh2306\}@msstate.edu}
\date{}

\begin{document}

    \maketitle
    \begin{abstract}
            Land use/land cover change (LULC) maps are integral resources in earth science and agricultural research. Due to the nature of such maps, the creation of LULC maps is often constrained by the time and human resources necessary to accurately annotate satellite imagery and remote sensing data. While computer vision models that perform semantic segmentation to create detailed labels from such data are not uncommon, little research has been done on self-supervised and unsupervised approaches to labelling LULC maps without the use of ground-truth masks. Here, we demonstrate a self-supervised method of land cover segmentation that has no need for high-quality ground truth labels. The proposed deep learning employs a frozen pre-trained ViT backbone transferred from DINO in a STEGO architecture and is fine-tuned using a custom dataset consisting of very high resolution (VHR) sattelite imagery. After only 10 epochs of fine-tuning, an accuracy of roughly 52\% was observed across 5 samples, signifying the feasibility of self-supervised models for the automated labelling of VHR LULC maps.
    \end{abstract}
    
    \begin{multicols}{2}
    \section{Introduction}

    High-quality land cover use and land cover change (LU/LC) maps are critical to understanding and evaluating human interaction with natural resources and the effects of such interactions on both the natural environment and humans. LU/LC maps are critical in documenting climate change, deforestation, wildland fires, and atmospheric research \cite{pielkeLandUseClimate2005, feddemaImportanceLandCoverChange2005, royGlobalAnalysisTemporal2021}. Numerous advancements in remote sensing equipment, computational resources, and artificial have somewhat alleviated the need for large scale human-annotation of satellite imagery and remote sensing data, yet little research on scaling such techniques to very high resolution (VHR) data is still in its infancy \cite{wulderLandCover2018,mohanrajanSurveyLandUse2020}. The goal of this work is to address the potential of self-supervised methods of training to overcome the labeled data bottleneck and to determine if the performance of such a model is applicable to the creation of deep learning techniques for the segmentation of land cover maps. By using a STEGO framework to distill correspondence similarities from a pre-trained model, our aim is to utilize existing models to determine the semantic information while evaluating feature correspondences in VHR satellite imagery to create detailed LULC maps.

    \section{Related Work}

        \subsection{Land Use/Land Cover Change Maps}

        Land use/land cover change (LU/LC) maps detail natural resource usage by a variety of researchers in various fields. LU/LC allow researchers to monitor developments in environmental interactions as well as model future changes in land use change \cite{feddemaImportanceLandCoverChange2005,zelekeImplicationsLandUse2001,pielkeLandUseLand2011}. However, development of high-quality large-scale land cover maps is labor-intensive and computationally expensive, and as such resolution is typically constrained: USGS NLCD 2019 maps land usage at a resolution of 30m. Due to rapid developments in remote sensing, the availability of very high resolution imagery and satellite data is now abundant, yet research on scaling algorithms used for labeling coarser data is still in its relative infancy \cite{homerConterminousUnitedStates2020}. With the emergence of recent self-supervised algorithms that possess the ability to learn semantically rich features without the need for high-quality annotations, further research is needed to evaluate their potential for labeling LULC maps at high resolution \cite{chenSimpleFrameworkContrastive2020,caronEmergingPropertiesSelfSupervised2021}. 

        \end{multicols}
        \begin{figure}[H]
            \centering
            \includegraphics[width=\textwidth]{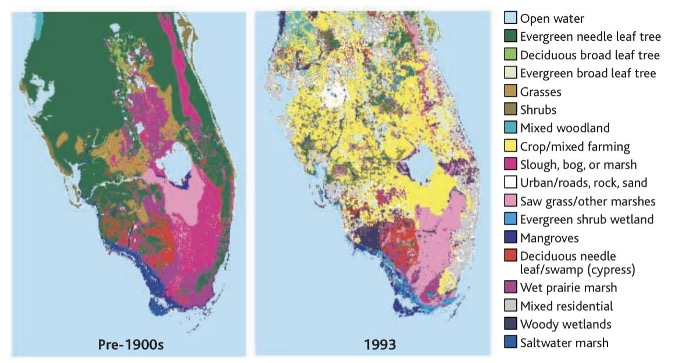}
            \caption{Example of large-scale LULC map of central and southern Florida, USA. LULC maps such as these are critical in evaluating and modeling human utilization of land and resources \cite{pielkeLandUseClimate2005}.}
            \label{fig:FLlandcover}
        \end{figure}
        \begin{multicols}{2}

        \subsection{Semantic Segmentation}

        Hamilton \textit{et al.} define semantic segmentation as “Semantic segmentation is the process of classifying each individual pixel of an image into a known ontology” \cite{hamilton2022unsupervised}. Semantic segmentation is useful when all objects or features in an image and their location(s) within the image is necessary information. To date, most deep-learning semantic segmentation algorithms employ the use of convolutional neural networks (CNNs) in some fashion; CNNs are a special class of deep learning models powered predominantly by stacked convolutional layers that enable neural networks the ability to better capture low-level, mid-level, and high-level spatial features \cite{heDeepResidualLearning2015, caronEmergingPropertiesSelfSupervised2021}. In remote sensing, research on the use deep learning architectures to label land use and land change has emerged, but most implementations rely on access to hand-labeled annotations specific to the region of investigation of the desired resolution, which is not always feasible \cite{nguyenMappingGlobalEcoenvironment2019,martinsDeepLearningHigh2022}.

        \subsection{Convoluional Neural Networks}
        
\begin{figure}[H]
            \centering
            \includegraphics[width=\textwidth]{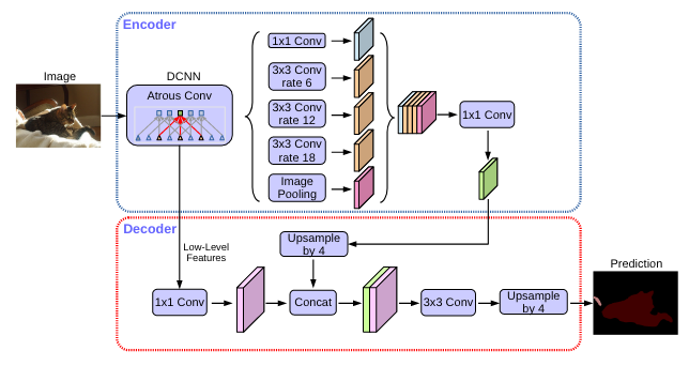}
            \caption{Simplified graph of DeepLabv3+ architecture. Note the extensive use of convolutional layers arranged in a hierarchy. Supervised architectures such as these perform well in scenarios where masked training data is of high-quality and abundant \cite{chenEncoderDecoderAtrousSeparable2018,chenRethinkingAtrousConvolution2017}.}
            \label{fig:deeplab}
        \end{figure}

        CNNs primarily consist of convolutional layers, pooling layers, and fully connected layers arranged in such a hierarchy to allow a representation of low-level features (edges, lines, corners) to be used to learn more complex high-level features (faces, objects, etc.) \cite{heDeepResidualLearning2015}. CNNs are widely popular in most computer vision applications, with tasks ranging from image classification to optical flow, and as such are commonly used in the processing of spatial remote sensing data \cite{martinsExploringMultiscaleObjectbased2020}. For semantic segmentation specifically, DeepLabv3+ (see figure \ref{fig:deeplab}) is one of the most popular frameworks for vision tasks that use supervised learning \cite{chenRethinkingAtrousConvolution2017,chenEncoderDecoderAtrousSeparable2018}.

        \end{multicols}
        \begin{multicols}{2}

        \subsection{Contrastive Learning}

        Contrastive learning is a framework for self-supervised learning that operates on the principle that representations of samples from similar classes should be similar, whereas representations of samples from disparate classes should be dissimilar. Many common contrastive learning algorithms enforce this principle with the use of data augmentation: a pair of subsamples is created by taking a single image as input (referred to as an "anchor") before randomly cropping, stretching, or resizing the image in such a way to create two dissimilar images that belong to the same class - otherwise known as a positive pair. Negative pairs consisting of two augmented images from different classes are also used in training with one exception; their representations should be dissimilar. A contrastive loss function attempts to maximize agreement between positive pairs and minimize agreement between negative pairs. Specifically, when classes are unknown, other augmented images in a given batch that are \textbf{not} derived from the original “anchor” image are treated as negative pairs. \cite{jaiswalSurveyContrastiveSelfsupervised2021, chenSimpleFrameworkContrastive2020}.

        \begin{figure}[H]
            \centering
            \includegraphics[width=0.5\textwidth]{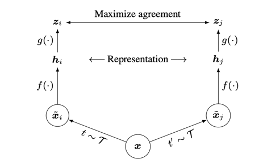}
            \caption{Graphical depiction of SimCLR - a popular contrastive learning framework. An input \(X\) is augmented into a positive pair. Loss is computed over the representations received from a multilayer perceptron (MLP) such to minimize the differences between the two representations during training \cite{chenSimpleFrameworkContrastive2020}.}
            \label{fig:clr}
        \end{figure}

        \subsection{Transformers in Computer Vision}

        Self-attention is an important component of state-of-the art models built for natural language processing (NLP). Implementing self-attention into computer vision models has the potential to allow deep learning models to learn features in an image in a global fashion rather than local representations confined to the receptive fields of convolutional layers. The most popular vision-based adaptations of these methods draw from Transformer-based models commonly applied in NLP domains such as BERT and GPT-3 with the goal of overcoming such limitations on learning long-range interactions \cite{fuVisionTransformerVit2022,liuSwinTransformerHierarchical2021,caronEmergingPropertiesSelfSupervised2021,dosovitskiyImageWorth16x162021,jaiswalSurveyContrastiveSelfsupervised2021}.

        Vision Transformer (ViT) is a family of self-attention-based architecture that divides an image into a series of patches that are then placed into a multi-head self attention block. Patches are flatted into a linear projection before input into a large model consisting of several stacked Transformer encoders - each consisting of global self-attention layers. Self-attention layers are then fed into multi-layer perceptrons (MLP) at the final stage of an encoding layer. Residual connections concatenate the input of the encoder to the output of the multi-head self attention block, and the output of said self attention blocks to the final output of the layer. Positional encodings are also inputted alongside each linearly embedded patch, but these encodings do not convey information regarding the position of each embedded patch before training - meaning spatial information and relationships must be learned from scratch (as opposed to CNNs). Regardless, the performance of ViTs is competitve with or exceeds that of state-of-the-art CNN methods for classification when trained on several benchmark datasets while being considerably more computationally efficient in terms of resource use during pre-training. However Big Transformer (BiT) - a family of similar architecture that utilizes large CNNs during pre-training - yielded comparatively better results when the number of pre-training samples was reduced \cite{dosovitskiyImageWorth16x162021}.

        \end{multicols}
        \begin{figure}[H]
            \centering
            \includegraphics[width=\textwidth]{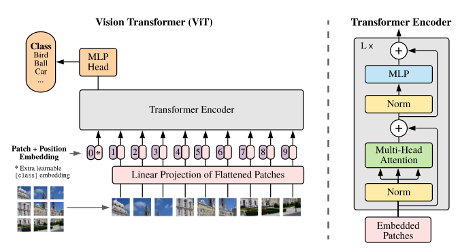}
            \caption{Graphical overview of basic ViT architecture. Note the multi-head self attention layer included in each transformer encoder layer \cite{dosovitskiyImageWorth16x162021}.}
            \label{fig:vit}
        \end{figure}
        \begin{multicols}{2}

        \subsection{DINO}
        Whilst ViTs have shown to be of similar or higher performance compared to state-of-the-art CNNs, their large size and need for huge amounts of labeled pre-training data make them relatively impractical for implementations in situations where computational resources and data availability are constraints. Caron \textit{et al.} showed that a self-supervised approach to pre-training using contrastive learning showed considerably better performance on downstream segmentation tasks compared to models pre-trained in a fully supervised procedure. This approach, named DINO (\textbf{Di}stilling knowledge with \textbf{no} labels) was able to learn features that were richer and more semantically meaningful compared to those learned by ViTs trained in a fully supervised fashion - making it suitable for usage in various vision tasks beyond classification. DINO’s ability to learn rich features from samples without annotation stems from its use of knowledge distillation to train a new model against an ensemble of previous models using augmented pairs from the same sample \cite{hinton2015distilling,caronEmergingPropertiesSelfSupervised2021}.

        \begin{figure}[H]
            \centering
            \includegraphics[width=0.5\textwidth]{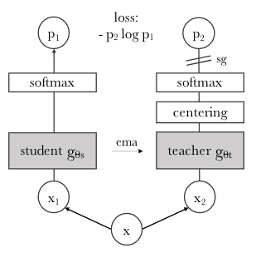}
            \caption{Graphical overview of DINO training. Note how the training step utilizes a contrastive learning approach. Knowledge distillation is used to train a teacher and student model, here student model \(g_{\theta s}\) (ViT backbone at current epoch) is trained to match the output of teacher model \(g_{\theta s}\) (ensemble of ViT backbones at all previous epoch). Using such a knowledge distillation during training was shown to yield consistent high-quality targets from the teacher model \cite{caronEmergingPropertiesSelfSupervised2021}.}
            \label{fig:dino}
        \end{figure}

        \subsection{STEGO}
        One key observation from DINO is that once features have been captured in a self-supervised fashion during pre-training, correlations between learned features are consistent with semantic information found not only within the same image but across other images as well \cite{caronEmergingPropertiesSelfSupervised2021}. STEGO is an architecture recently introduced to perform unsupervised semantic segmentation by distilling feature correspondences (retrieved from a frozen pre-trained DINO ViT backbone) across and between samples into a lower dimensional representation using a simple MLP. It requires no fine-tuning due to DINO's ability to capture rich features in pre-training - the outputs of the frozen DINO backbone are fed to a single segmentation head consisting of a simple MLP. By utilizing a basic but creative contrastive learning implementation where a contrastive loss function takes into account the feature correspondences in a pair of samples, this method has yielded breakthrough results on unsupervised semantic segmentation tasks. Further, the architecture is highly adaptable to various scenarios and domains as the architecture itself does not define a strict method for pre-training, thus any future advancements in vision models or pre-training paradigms can be easily patched in \cite{hamilton2022unsupervised}.

        \end{multicols}
            \begin{figure}[H]
                \centering
                \includegraphics[width=\textwidth]{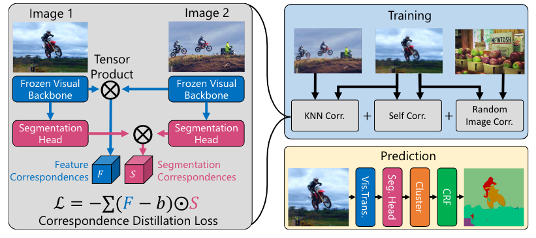}
                \caption{Graphical overview of STEGO model. Note the  \cite{hamilton2022unsupervised}.}
                \label{fig:stego}
            \end{figure}
        \begin{multicols}{2}

    \section{Materials \& Methodology}

        \subsection{Datasets}

            For this research, two datasets consisting of VHR satellite imagery and their corresponding masks were used in various capacities in training and evaluation. Whilst pre-training was done using DINO with data from ImageNet, parameters in the backbone ViT were transferred to the downstream STEGO model for fine-tuning. Ideally, a sufficiently large dataset of VHR imagery would be used to pre-train DINO whereas a smaller dataset would then be used to fine-tube. To the authors' knowledge, no such publicly accessible dataset exists at this point in time.

            \subsubsection{DeepGlobe}

            The DeepGlobe dataset is a high-resolution satellite image dataset that consists of three challenge tracks: road extraction, Building Detection, and Land Cover Classification. We utilized the Land Cover Classification track, which is in accordance with the aim of this project. Land cover classification is important for tasks such as: sustainable development, autonomous agriculture, and urban planning. The Land Cover Classification task is split into seven classes: urban, agriculture, rangeland, forest, water, barren, and unknown. This is defined as a multi-class segmentation task. 

            The DeepGlobe Land Cover Classification consists of 1,146 images at a spatial resolution of 50 centimeters per pixel. Inputs were resized to 256x256x3 with a resultant spatial resolution of 5 meters per pixel \cite{demirDeepGlobe2018Challenge2018}.

            \begin{figure}[H]
                \centering
                \includegraphics[width=0.5\textwidth]{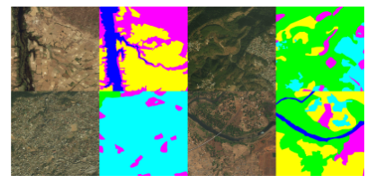}
                \caption{Selected images and masks from DeepGlobe Land Cover Challenge Dataset \cite{demirDeepGlobe2018Challenge2018}.}
                \label{fig:deepglobe}
            \end{figure}

            \subsubsection{LandCover.ai}

            Similar to the DeepGlobe dataset (Land Cover Classification task), the LandCover.ai dataset is an RGB manually-annotated image dataset based on satellite images for land cover classification. The images from this dataset come from Central Europe with most images coming from Poland. The landscape of Poland is primarily dominated by mixed forests and agrarian areas. This dataset is divided into four different classes: buildings, woodlands, water, and road for land cover generality and usefulness for public administrations. However, we discarded the need for labels from this dataset as we only used the source images to train the STEGO model.

            In the Land Cover Classification track, the labels are only publicly available for the train splits. We decided to combine the validation and test source images, along with images from the LandCover.ai dataset, to create a pseudo-train dataset. This allows us to obtain accuracy values from the DeepGlobe train set. 

            The LandCover dataset consists of 10,604 images with spatial resolutions ranging from 50 centimeters per pixel to 25 centimeters per pixel. After resizing to 256x256x3, the resultant spatial resolution ranged from 50-100 centimeters per pixel \cite{boguszewskiLandCoverAiDataset2022}.

            \begin{figure}[H]
                \centering
                \includegraphics[width=0.5\textwidth]{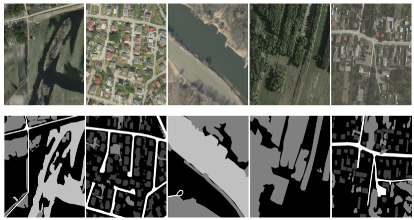}
                \caption{Selected images and masks from LandCover.ai dataset \cite{boguszewskiLandCoverAiDataset2022}.}
                \label{fig:landcoverai}
            \end{figure}
        
        \subsection{Model Configuration}

        In order to achieve best results, we opted to utilize various vision transformers that were pre-trained by the DINO paradigm. It is important to note that the vision transformers were pre-trained on the ImageNet dataset. This dataset includes 14+ million images. This allows the vision transformer to see a diverse array of pre-training data and allows the potential for higher segmentation accuracy compared to models pre-trained on smaller datasets \cite{caronEmergingPropertiesSelfSupervised2021}. In order to train the STEGO segmentation decoder, we utilized a set batch size of 16, learning rate of 0.0001, an Adam optimizer, and momentum constraints preset by the STEGO authors. It is also important to note that the learning rate is scalable, as the STEGO architecture allows for such a mechanism \cite{hamilton2022unsupervised}. We set training and inference images to a set size of 256 x 256 to allow for the model to train efficiently while relaxing memory constraints and also prevent the loss of valuable global information. We limit the model to train on ~200 images with around 10 epochs to train due to compute oonstraints.

    \section{Results}

\begin{figure}[H]
            \centering
            \includegraphics[width=\textwidth]{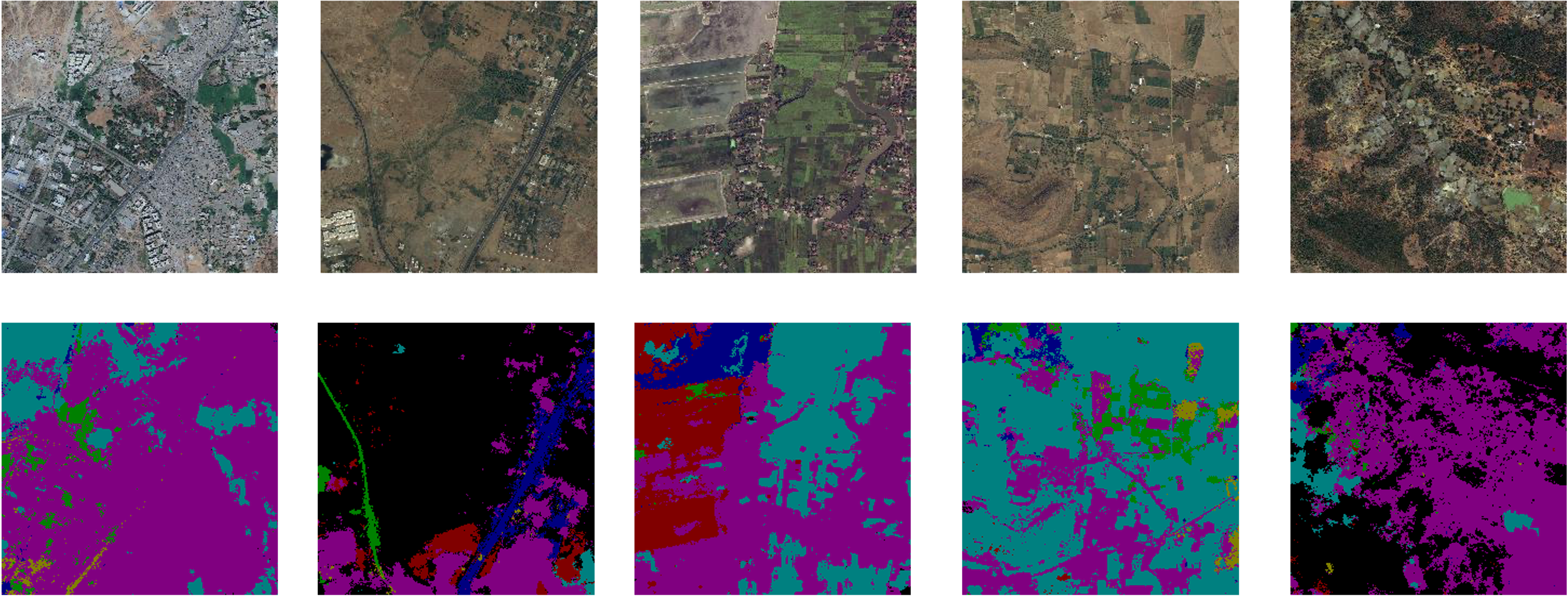}
            \caption{Images segmented using proposed model. Images on the top were segmented using the STEGO model after fine-tuning, masks on the bottom were the resultant masks from the segmentation component of the model. Despite only 10 epochs of fine-tuning and no ground truth labels to rely on, the model was still able to discern differences in spatial features to yield a fairly accurate map of land cover.}
            \label{fig:stego_results}
        \end{figure}

        Given the limited number of epochs and small training dataset, the results are greater than expected. However, we expect the quality of the features from the vision transformer and the possibility of the model reaching an optima quickly to be the reasoning behind this. In most, if not all, of the inference images, STEGO performs well in identifying segmentation areas, but does seem to over-represent some classes, resulting in a class imbalance. We were able to observe the model achieving a ~52\% accuracy across 5 test samples with the given model hyperparameters (see figure \ref{fig:stego_results}). We believe that with training data and better optimized hyperparameters, a modified STEGO model can achieve state-of-the-art performance while simply utilizing a pre-trained model from DINO. It is important to note that this result was achieved without a domain focused pre-trained dataset. We can expect the model to perform better using the weights from a DINO backbone pre-trained with a sufficiently large dataset of consistent high-resolution satellite imagery - such a dataset is not publically available to the authors' knowledge.

        \end{multicols}
        \begin{multicols}{2}

    \section{Discussion}
    
    We found that utilizing a pre-trained DINO vision transformer as a backbone for the STEGO decoder can yield decent results in semantic segmentation. We were able to obtain ~52\% accuracy with a small batch size and minimal training images. Considering that accuracy results over 70\% are typically considered acceptable for models that utilize fully supervised training in the remote sensing research domain \cite{chenMFANetMultiLevelFeature2021, karraGlobalLandUse2021, gibrilIntegrativeImageSegmentation2018}, these results are promising. We believe that scaling up the datasets used in both training phases will not only yield better results but will also improve the model's robustness. Although STEGO is proven to be a good option for semantic segmentation, there exist potential improvements to the architecture and methodology. One such potential modification would be to incorporate the MASK DINO framework into the STEGO architecture to take advantage of potentially richer learned semantic features \cite{liMaskDINOUnified2022}. Another potential investigation would be the performance of the MoBY architecture with the improved SwinV2 Vision Transformer variant on VHR satellite imagery segmentation tasks due to SwinV2's ability to handle high-resolution samples \cite{xieSelfSupervisedLearningSwin2021,liuSwinTransformerV22022}. In addition to LULC segmentation, there are various similar domains where such work can be applied to. Large scale terrain segmentation is an important area of study in relation to land cover classification. An accurate model for terrain segmentation can easily assist tasks in robotics such as exploration. This provides a reason to scale up from aerial images to land masses and remains a widely researched topic in computer vision. Another area of investigation that stems from this project is the concept of using real world data and observations to fix model inaccuracies during a robotics task such as exploration or path planning. Successfully incorporating real world data could have the effect of needing less training data to build models, thus saving time and resources training models.

    \section{Conclusion}
    Land cover classification is still an important problem to date, and is an important task in the field of computer science for agricultural and commercial reasons \cite{demirDeepGlobe2018Challenge2018, giriNextGenerationGlobal2013}. In this study, we have shown that the combination of DINO and STEGO has the potential to be a feasible alternative to supervised learning models. Contrastive learning is relatively new in the area of computer vision but has the ability to generate robust models that can be applied to various computer vision tasks: such models can provide many benefits, and we believe that smaller models trained with contrastive learning can achieve performance competetive with supervised learning methods. Self-supervised learning offers a good alternative for supervised training on unreliable training datasets \cite{xieSelfSupervisedLearningSwin2021, jaiswalSurveyContrastiveSelfsupervised2021}, and we have shown that self supervision can do an acceptable job at distinguishing different land cover classes in VHR aerial images. As such, further research could potentially reveal solutions to problems in earth science, climatology, and agriculture.

    \end{multicols}

    \section{References}

    \bibliographystyle{IEEEtran}
    \bibliography{references}

\end{document}